%% file: grace.tex
\documentclass[12pt]{article}

\input preamble.tex

\title{Detecting AI-generated Artwork}
\author{Meien Li and Mark Stamp}

\author{Meien Li\footnotemark[1]\,\,\footnotemark[2]\ \ \ 
Mark Stamp\footnotemark[1]}

\begin{document}

\symbolfootnotetext[1]{Department of Computer Science, San Jose State University}
\symbolfootnotetext[2]{meien.li$@$sjsu.edu}

\maketitle

\begin{abstract}
The high efficiency and quality of artwork generated by Artificial Intelligence (AI)
has created new concerns and challenges for human artists.  
In particular, recent improvements in generative AI have made it difficult for people 
to distinguish between human-generated and AI-generated art. In this research, 
we consider the potential utility of various types of Machine Learning (ML)
and Deep Learning (DL) models in distinguishing AI-generated artwork from 
human-generated artwork. We focus on three challenging artistic styles, 
namely, baroque, cubism, and expressionism. The learning models we test are 
Logistic Regression (LR), Support Vector Machine (SVM), Multilayer Perceptron (MLP), 
and Convolutional Neural Network (CNN). Our best experimental results yield
a multiclass accuracy  of~0.8208 over six classes,
and an impressive accuracy of~0.9758 for the binary classification 
problem of distinguishing AI-generated from human-generated art. 
\end{abstract}

\section{Introduction}

Artificial Intelligence (AI) has achieved significant advances in recent 
years~\cite{huang2022}. AI-related technologies involving 
Machine Learning (ML) and Deep Learning (DL) have resulted in many successes
and the concept of generative AI has become well-known to the public through the 
existence of interactive AI models, including chatbots and image generators. 
However, AI-generated artwork raises concerns for human artists. For example, 
human-generated art is used to train AI models without permission from---or compensation 
to---human artists. It has also becomes difficult for humans to distinguish between 
human-generated and AI-generated art, and it is therefore important to investigate
new approaches to assist humans in detecting AI-generated artwork. 

The goal of the research presented in this paper is to investigate the performance 
of various types of classic ML models and compare their performance to DL models 
for distinguishing human-generated from AI-generated art. 
We consider both binary and multiclass classification problems, 
based on a challenging image dataset that we have collected. 
As part of this research, we also consider various
features and feature reduction techniques. 

The remainder of the paper is organized as follows. 
Section~\ref{sect:background} gives relevant background information,
while Section~\ref{sect:dataset} provides an overview of our dataset
and we discuss the features that we extract. 
Section~\ref{sect:experiments} discusses the experiments that we perform
and the results that we obtain. Finally, in Section~\ref{sect:conclusion},
we provide concluding remarks, and we briefly discuss future research directions.

\section{Background}\label{sect:background}

In this section, we discuss background topics that are relevant to
later sections of this paper. This background includes a brief discussion of AI-generated artwork and an introduction to the various ML and DL
models that are used in this research. 

\subsection{AI Generated Artwork}

Due to recent advances in generative AI, tools to generate high quality
AI art are readily available to the public. Many people have become ``prompt artists'' 
by using online AI models such as DALL\_E~\cite{dalle2022} and 
Stable Diffusion~\cite{zhou2023} to generate artwork---entering a simple text 
description of the desired art is all that is required
of the ``artist''. Moreover, the application programming interface (API) is available. 
online for tools such as the DeepAI~\cite{deepai2024} image generator, 
further expand the possibilities for more advanced users. 

AI-generated art has achieved a high level of quality,
and researchers have investigated the application of such artwork 
to a variety of disciplines. Examples of applications of AI-generated art include 
illustrations in anatomical science~\cite{noel2024} and visual arts 
education~\cite{dehouche2023}. 
However, there is also considerable controversy regarding AI-generated art. 
For example, the general availability of AI image generators 
raises concerns regarding the ownership 
of the resulting artwork: Should the artwork belong to the AI model, 
the designers of the model, the users of the model, the artists whose 
artwork was involved in training the model, or some combination thereof? 
Besides, social media platforms such as $\mathbb{X}$ (formerly known as Twitter) 
have updated their terms of service so that they can use posted artwork 
to train AI models. 

The quality of generative models has reached the point where
people often cannot distinguish between human-generated and
 AI-generated art~\cite{samo2023}. This indicates the importance of 
determining whether tools can be developed to assist humans in 
distinguishing between human-generated and AI-generated artwork. 
Based on the performance of learning model for image classification, 
in this paper we consider the utility of such models for distinguishing between 
human-generated and AI-generated artwork. 
Specifically, we consider Logistic regression (LR), Support Vector Machine (SVM),
and Multilayer Perceptron (MLP) models trained on extracted features,
and we compare these results to Convolutional Neural Network (CNN) 
models that are trained directly on images.
We consider both binary classification (i.e., distinguishing human-generated
from AI-generated art) and multiclass classification (i.e., distinguishing between 
the six categories of art in our dataset).

\subsection{Learning Models}

In this section, we will introduce the models that we will employ in our experiments. 
These models are 
Logistic Regression (LR), 
Support Vector Machine (SVM), 
Multilayer Perception (MLP), 
and Convolutional Neural Network (CNN). 

\subsubsection{Overview of Logistic Regression}

Logistic Regression (LR) is a classic supervised ML model that can be used for
classification problems. The basic idea behind this model is to use a logistic 
function to transform input data into a probability (i.e., a value in the interval 
between~0 and~1), which indicates the likelihood that the given data correspond 
to a specified category. Although LR model is relatively simple, it often perform 
well in practice~\cite{agrawal2023}.

\subsubsection{Overview of Support Vector Machine}

Support Vector Machine (SVM) is a supervised ML model that is widely used for
classification. The core idea of SVM is to construct a hyperplane that distinguishes 
different labeled data, with the goal of maximizing the margin between the classes, 
The use of various kernel functions allows for nonlinear decision boundaries~\cite{stamp2022}.

\subsubsection{Overview of Multilayer Perception}

Multilayer Perceptrons (MLP) is a DL model that is also widely used for supervised 
learning tasks. A perceptron acts as a simulated neuron structure~\cite{zhao2021}. 
An MLP model consists of multiple layers of interconnected perceptrons,
with the output of the previous layer passed on as the input of the next layer.
An MLP is somewhat analogous to an SVM where the equivalent of the kernel
function is learned, rather than specified by the user~\cite{stamp2022}.

\subsubsection{Overview of Convolutional Neural Networks}

Convolutional Neural Network (CNN) is another DL model that is closely
associated with image analysis~\cite{lecun2015deep}. CNN models include 
convolutional layers, pooling layers, and at least one dense layer,
and they are designed to efficiently deal with local structure.

\section{Dataset and Features}\label{sect:dataset}

In this section, we first introduce our image dataset. Then we summarize the 
features that we extract from the image dataset, which are used to train our
LR, SVM, and MLP models.

\subsection{Image Dataset}

To test our classification models, we want to consider artistic styles that can be challenging 
to distinguish. Therefore, we choose baroque, cubism, and expressionism as the target 
styles, with human-generated and AI-generated examples in each of these styles.

\begin{itemize}
\item Baroque {---} The word baroque is thought to derive from 
``misshapen and irregular pearls''~\cite{sun2021}. 
As an artistic style, baroque focuses on a mixture of reality and fiction 
with dramatic contrasts of color, light, and shadow~\cite{sun2021}. 
This style has more irregular curves and dynamics compared to artwork of
the Renaissance period, yet it follows the basic principles of perspective and realism.  
    
\item Cubism {---} Cubism is considered to be an example 
of experimental art, that is, art that breaks 
traditional concepts of aesthetics~\cite{gombrich1995}. 
Many cubists intend to extract elements 
from an observation and break regular objects that are then rearranged 
and combined. Cubists reject observing the world from a fixed perspective~\cite{sun2021},
and cubist art is full of abstract shapes. 
    
\item Expressionism {---} Like cubism, expressionism is considered to be an
experimental style of art, but expressionism focuses on 
distorting nature~\cite{gombrich1995}. Expressionists relies on emotional 
and sensual expression that emphasizes the spiritual world~\cite{sun2021}. 
Since artists discard subject-matter and choose the creation of 
abstract, expressionism shares some 
characteristics with cubism~\cite{gombrich1995}. 
\end{itemize}

To obtain a sufficiently large dataset that contains both human-generated 
and AI-generated artwork, we have combined images from the 
WikiArt~\cite{wikiart2024} and AI-ArtBench~\cite{silva2023} datasets, 
which are available from Kaggle~\cite{elkholy2024}. 
With this combined dataset, we obtain samples for five of the six classes
that we will use in our experiments, with only the AI-generated cubism images missing, 
We use the DeepAI~\cite{deepai2024} image generator---an AI image generator available online---to 
create our AI-generated cubism samples. 

In summary, our dataset consists of human-generated artwork from 
the WikiArt dataset, while for our AI-generated artwork, the samples of both 
baroque and expressionism styles artwork were obtained from the 
AI-ArtBench dataset and generated with 
Latent Diffusion (LD)~\cite{rombach2022} model, 
with the AI-generated cubism samples 
generated using DeepAI. 
We summarize our dataset in Table~\ref{dataset}. 

\begin{figure}[!htb]
    \centering
    \begin{tabular}{ccc}
    \includegraphics[width=0.24\textwidth]{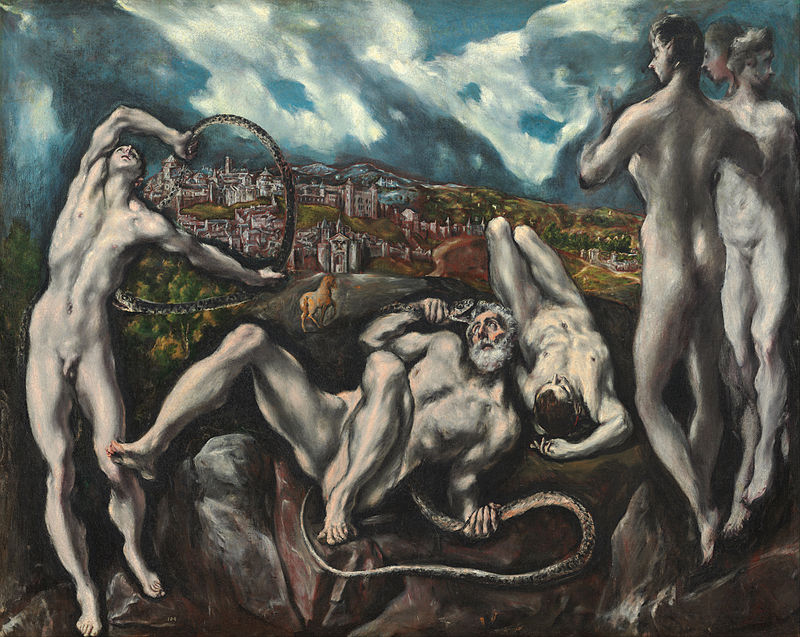}
    &
    \includegraphics[width=0.24\textwidth]{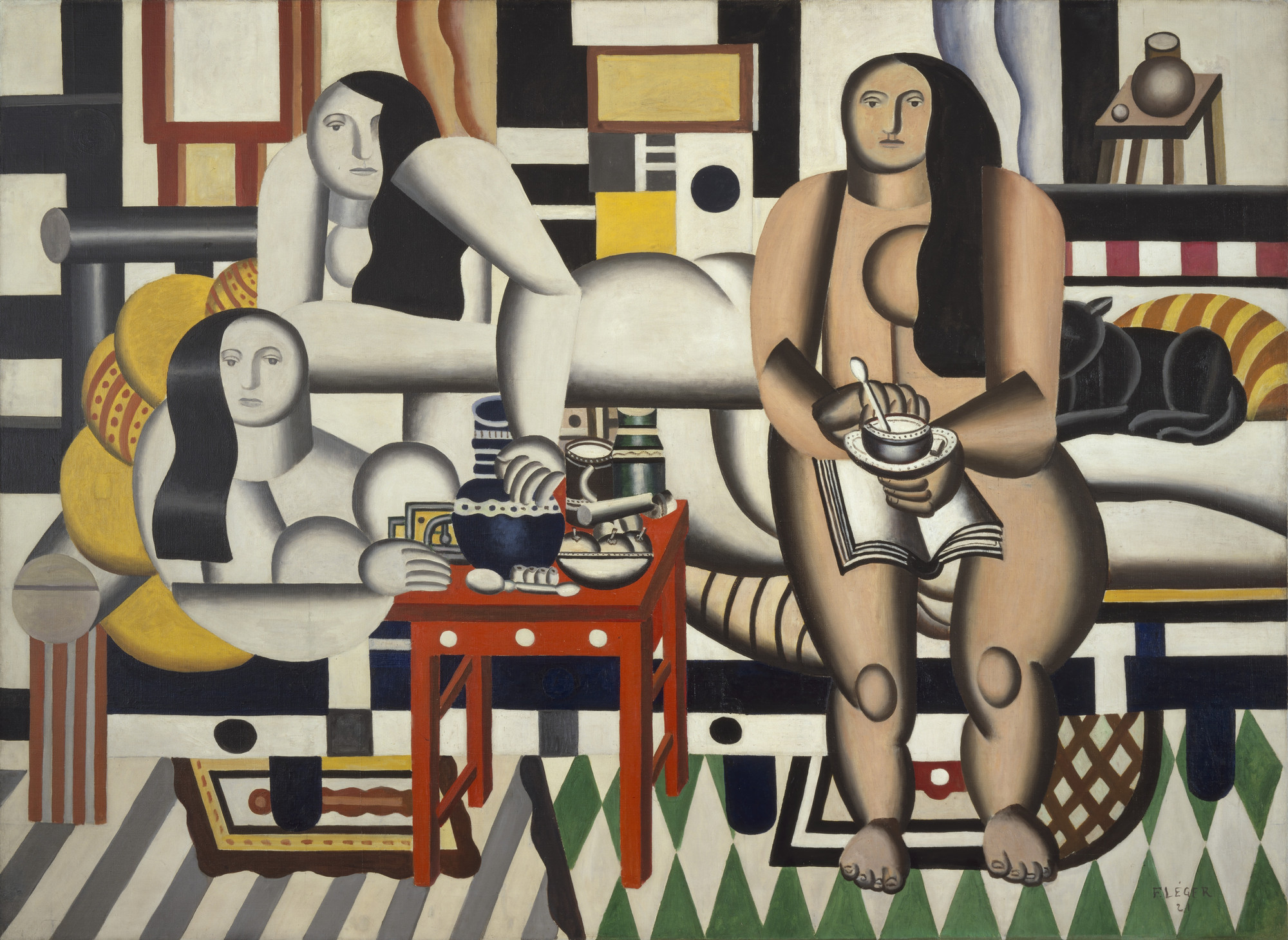} 
    &
    \includegraphics[width=0.24\textwidth]{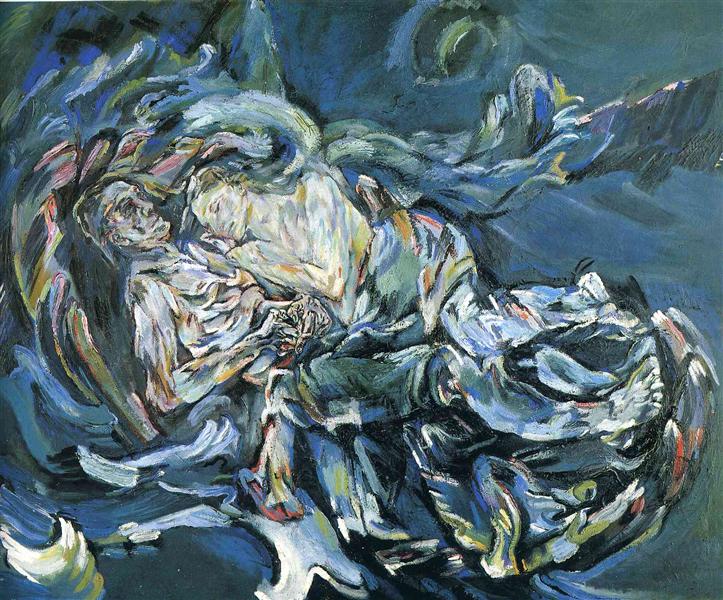}
    \\
    (a) Baroque
    &
    (b) Cubism
    &
    (c) Expressionism
    \end{tabular}
    \caption{Examples of human-generated art}\label{fig:humanArt}
\end{figure}

In Figure~\ref{fig:humanArt}, we give examples of human-generated art
in the styles of Baroque, Cubism, and Expressionism.
Specifically, Figures~\ref{fig:humanArt}(a)--(c) are \textit{Laoco\"{o}n} by El Greco,
\textit{Three Women} by Fernand L\'{e}ger, and
\textit{The Bride of the Wind} by Oskar Kokoschka, respectively.
These samples were obtained from the WikiArt dataset.

\begin{figure}[!htb]
    \centering
    \begin{tabular}{ccc}
    \includegraphics[width=0.24\textwidth]{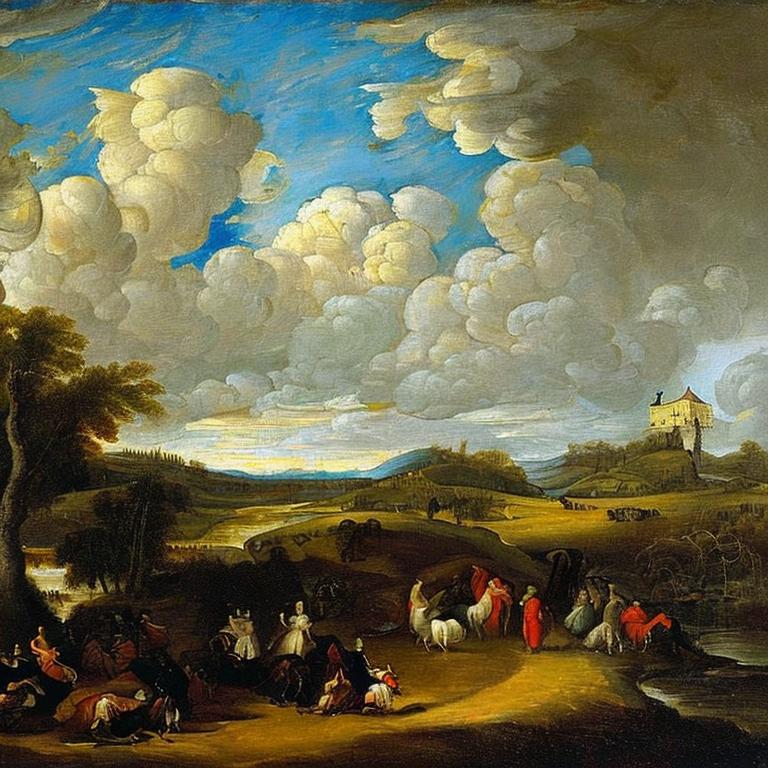}
    &
    \includegraphics[width=0.24\textwidth]{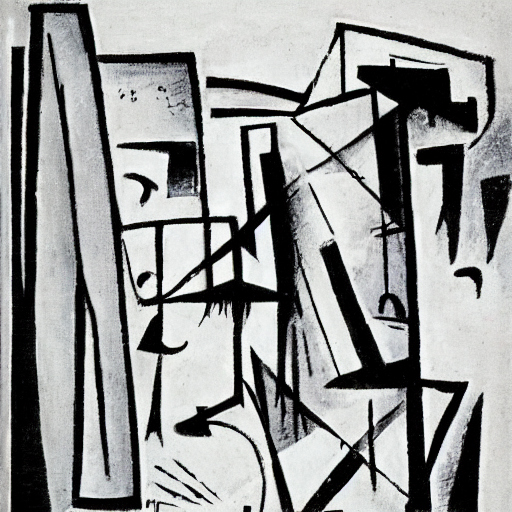} 
    &
    \includegraphics[width=0.24\textwidth]{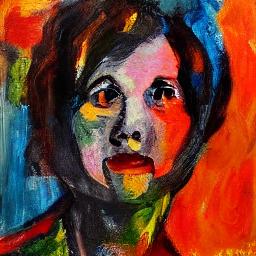}
    \\
    (a) Baroque
    &
    (b) Cubism
    &
    (c) Expressionism
    \end{tabular}
    \caption{Examples of AI-generated art}\label{fig:aiArt}
\end{figure}

In Figures~\ref{fig:aiArt}(a)--(c), we give examples of AI-generated art in the
Baroque, Cubism, and Expressionism styles, respectively.
Figures~\ref{fig:aiArt}(a) and (c) are samples from the 
AI-ArtBench LD dataset, while Figure~\ref{fig:aiArt}(b) was 
generated with the DeepAI image generator. 

\begin{table}[!htb]
\caption{Dataset}\label{dataset}
\centering
\adjustbox{scale=0.90}{
\begin{tabular}{c|c|c} 
\midrule\midrule
Category & Source & Samples \\
\midrule 
 Human-baroque & WikiArt & 1000 \\ 
 Human-cubism & WikiArt & 1000 \\ 
 Human-expressionism & WikiArt & 1000 \\ 
 AI-baroque & AI-ArtBench LD & 1000 \\ 
 AI-cubism & DeepAI & 1000 \\ 
 AI-expressionism & AI-ArtBench LD & 1000 \\ 
\midrule\midrule
\end{tabular}
}
\end{table}

\subsection{Features}

To extract features from the image dataset, we first resize all of the images to 
be~$255 \times 255$ pixels. We extract~39 features in total,
and these features can be broadly categorized as 
brightness, color, texture, shape, and noise~\cite{chavda2018}.
\begin{itemize}
\item Brightness features {---} We use the mean and entropy to quantify 
pixel brightness. 
\item Color features {---} We analyze the artwork images in the context of 
RGB (red, blue, green) and HSV (hue, saturation, and value) histograms. 
We quantify each of the R, G, and~B histograms by computing
mean, variance, kurtosis, and skewness, as well as the overall entropy. 
This results in~13 features derived from the RGB histograms. 
For each of the H, S, and~V histograms, we compute the 
variance, kurtosis, and skewness, as well as the overall entropy, 
for a total of~10 features from the HSV histograms.
\item Textual features {---} We use two approaches to quantify textual 
features. The Local Binary Pattern (LBP) captures similarities of each pixel to 
neighboring pixels~\cite{pietikainen2010}, while the 
Gray Level Co-Occurrence Matrix (GLCM)~\cite{sebastian2012} 
is another classic method for textural analysis. 
\item Shape features {---} Histogram of Oriented Gradients (HOG) and edgelen are 
typical shape features. HOG captures the intensity of
gradient changes~\cite{chavda2018}, while edgelen highlights the boundaries 
of objects in an image by using the Canny edge detector~\cite{canny1986}. 
We compute the mean, variance, 
kurtosis, skewness, and entropy to quantify the HOG. 
\item Noise features {---} We consider two noise features, namely,
the entropy of noise and Signal to Noise Ratio (SNR). 

\end{itemize}
A summary of these features is given in Table~\ref{featureDataset}. 
We provide histograms of selected features in Figure~\ref{fig:histos}
in the Appendix.

\begin{table}[!htb]
\caption{Data features}\label{featureDataset}
\centering
\adjustbox{scale=0.90}{
\begin{tabular}{c|c}
    \midrule\midrule
    Feature type (number) & Feature name \\
    \midrule
    \multirow{2}{*}{Brightness (2)} & mean\_brightness \\
                                & entropy\_brightness \\
    \midrule
    \multirow{9}{*}{Color (23)} & red/green/blue\_mean \\
                           & red/green/blue\_variance \\
                           & red/green/blue\_kurtosis \\
                           & red/green/blue\_skewness \\
                           & rgb\_entropy \\
                           & hue/saturation/value\_variance \\
                           & hue/saturation/value\_kurtosis \\
                           & hue/saturation/value\_skewness \\
                           & hsv\_entropy \\
    \midrule
    \multirow{6}{*}{Texture (6)} & contrast \\
                             & correlation \\
                             & energy \\
                             & homogeneity \\
                             & lbp\_entropy \\
                             & lbp\_variance \\
    \midrule
    \multirow{6}{*}{Shape (6)} & hog\_mean \\
                           & hog\_variance \\
                           & hog\_kurtosis \\
                           & hog\_skewness \\
                           & hog\_entropy \\
                           & edgelen \\
    \midrule
    \multirow{2}{*}{Noise (2)} & noise\_entropy \\
                           & snr \\
    \midrule\midrule
\end{tabular}
}
\end{table}

\section{Experiments}\label{sect:experiments}

In this section, we present our experiments results, and we discuss our findings.
First, we train LR, SVM, and MLP models using all of the
features discussed in the previous section. Then we experiment with
feature reduction for each of these same three models. In contrast, we
train CNN models directly on the images. 
In each case, we consider both binary classification experiments and 
multiclass experiments. For binary classification, we simply 
distinguish human-generated from AI-generated art, whereas 
the multiclass experiments consider all six classes in our dataset,
namely, human-baroque, human-cubism, human-expressionism,
AI-baroque, AI-cubism, and AI-expressionism.

As is standard practice in machine learning, we 
use the \texttt{StandardScaler} from \texttt{sklearn.preprocessing} 
to preprocess each of the~39 features
listed in Table~\ref{featureDataset} by subtracting 
the mean and scaling our data to unit variance. 
Furthermore, in all LR, SVM, and MLP experiments, 
we use an~80:20 training:test stratified split and 5-fold cross validation. 
For our CNN experiments, we use an~80:10:10 training:test:validation split
to deal with potential overfitting issues.

\subsection{Models Trained on All Features}\label{subsec:original}

In this section, we give results for LR, SVM, and MLP models trained using all~39 features.
We apply each of these models to both the binary and multiclass cases. 
In Section~\ref{sect:feature}, below, we consider the same set of experiments, but
based on reduced sets of features.

\subsubsection{LR Results}

For our LR models, we use \texttt{LogisticRegression} from \texttt{sklearn.linear\_model} 
to train and test. The hyperparameters tested are listed in Table~\ref{LRresult}, 
with the best hyperparameters for the binary classification problem 
given in \textbf{boldface}, the best for the multiclass 
case appearing in \textit{italics}, and any that are best for
both cases are in \textit{\textbf{italicized boldface}}. 
For LR models trained on all~39 features, 
the best binary classification model achieves~0.9275 accuracy, 
while we achieve an accuracy of~0.7725 for the multiclass case. 

\begin{table}[!htb]
\caption{LR hyperparameters tested}\label{LRresult}
\centering
\adjustbox{scale=0.90}{
\begin{tabular}{c|c} 
\midrule\midrule
Hyperparameter & Values tested \\
\midrule
 C (regularization) & 0.2, 0.3, 0.5, 0.7, 0.8, \textit{\textbf{1}} \\ 
 solver & \textit{lbfgs}, saga, \textbf{liblinear} \\
 penalty & \textit{\textbf{l2}}, elastincnet \\
 max\_iter & \textbf{50}, 80, \textit{100}, 120, 200, 500, 1000 \\
\midrule\midrule
\end{tabular}
}
\end{table}

\subsubsection{SVM Results}

We use \texttt{SVC} from \texttt{sklearn.svm} to train and test our SVM models. 
The hyperparameters tested are shown in Table~\ref{svmResult}, 
with the best for the binary classification problem
given in \textbf{bold} font, the best for the multiclass 
case appearing in \textit{italics}, and any that are best for
both cases are in \textit{\textbf{italicized boldface}}. 
For SVM models trained on all~39 features, 
the highest binary classification accuracy we achieve is~0.9742, 
and we achieve a best accuracy of~0.8041 for the multiclass classification problem.

\begin{table}[!htb]
\caption{SVM hyperparameters tested}\label{svmResult}
\centering
\adjustbox{scale=0.90}{
\begin{tabular}{c|c} 
 \midrule\midrule
Hyperparameter & Values tested \\
\midrule
 C (regularization) & 0.1, 1, \textit{\textbf{10}} \\ 
 gamma & 0.1, 1, 10, \textit{scale}, \textbf{auto} \\
 kernel & linear, \textit{\textbf{rbf}} \\
\midrule\midrule
\end{tabular}
}
\end{table}

\subsubsection{MLP Results}

We use \texttt{MLPClassifier} from \texttt{sklearn.neural\_network} to train and test
our MLP models. Table~\ref{mlpResult} presents the hyperparameters tested, 
with the best for the binary classification problem 
given in \textbf{bold} font, the best performing for the multiclass 
case appearing in \textit{italics}, while any that are best for
both cases are given in \textit{\textbf{italicized boldface}}. 
The highest accuracy achieved for the binary classification problem
is~0.9758, and the best case, we attain an accuracy of~0.8125 for 
the multiclass classification problem. 

\begin{table}[!htb]
\caption{MLP hyperparameters tested}\label{mlpResult}
\centering
\adjustbox{scale=0.90}{
\begin{tabular}{c|c} 
\midrule\midrule
Hyperparameter & Values tested \\
\midrule
 hidden\_layer\_sizes & (50,), \textit{(100,)}, \textbf{(50, 50)} \\ 
 activation & identity, logistic, \textit{\textbf{relu}} \\
 alpha & \textit{0.0001}, \textbf{0.05} \\
 random\_state & \textit{30}, \textbf{40}, 50 \\
 solver & \textit{\textbf{adam}} \\
 learning\_rate\_init & \textit{\textbf{0.0001}} \\
 max\_iter & 200, 300, \textit{\textbf{1000}} \\
\midrule\midrule
\end{tabular}
}
\end{table}

\subsection{Feature Elimination}\label{sect:feature}

The technique that we use is Recursive Feature Elimination (RFE),
in which we select subsets of features and train our models on the reduced feature set. 
Since we have a large number of features, it is important to consider feature reduction,
which will improve efficiency, and might also improve classification accuracy.
For each model discussed in this section, we use the same software and
hyperparameters as given in Section~\ref{subsec:original}, above.

\subsubsection{Reduced Feature LR Results}

\begin{figure}[!htb]
    \centering
    \begin{tabular}{cc}
    \includegraphics[width=0.45\textwidth]{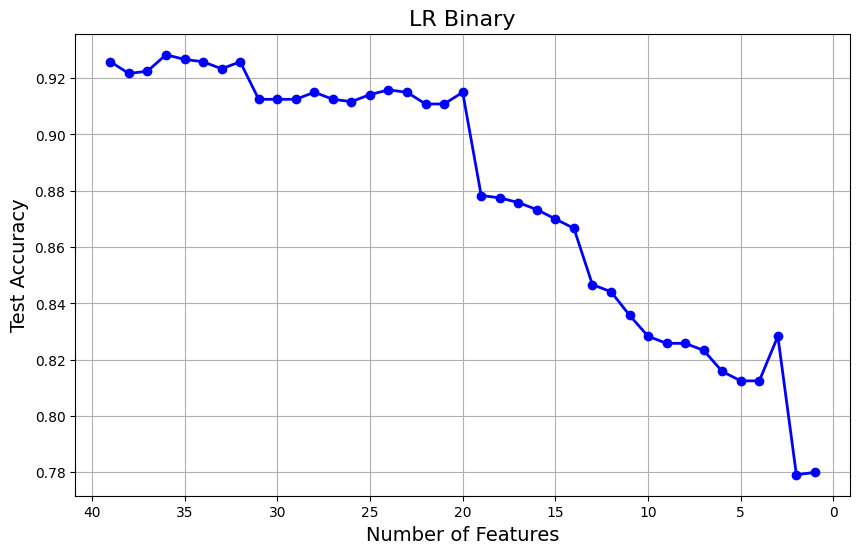}
    &
    \includegraphics[width=0.45\textwidth]{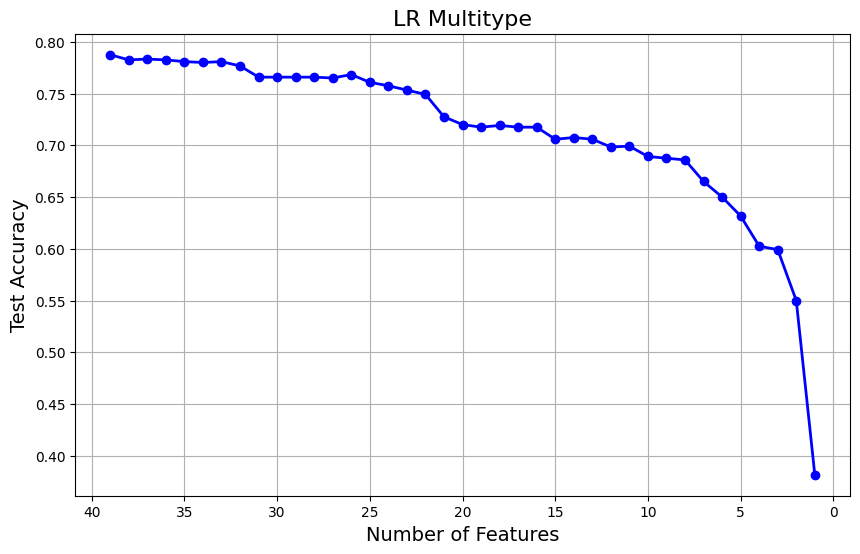}
    \\
    \adjustbox{scale=0.90}{(a) LR binary}
    &
    \adjustbox{scale=0.90}{(b) LR multiclass}
    \\ \\[-1.25ex]
    \includegraphics[width=0.45\textwidth]{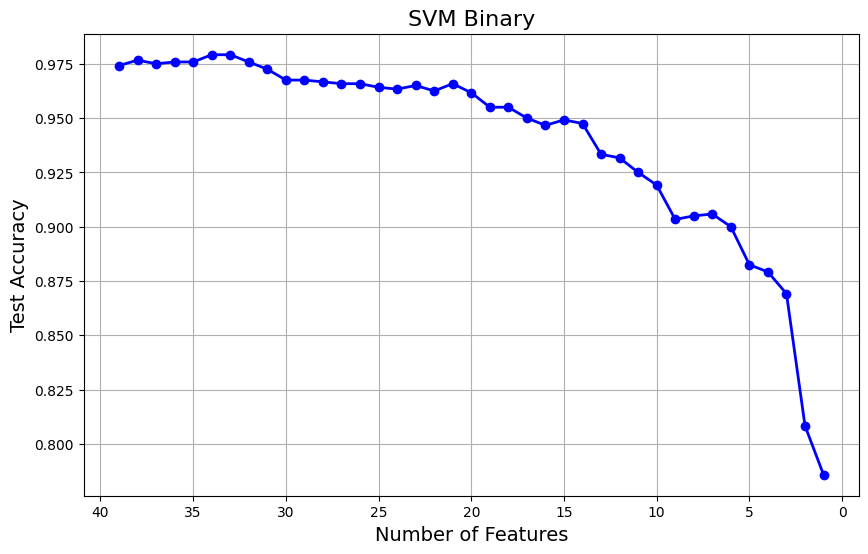}
    &
    \includegraphics[width=0.45\textwidth]{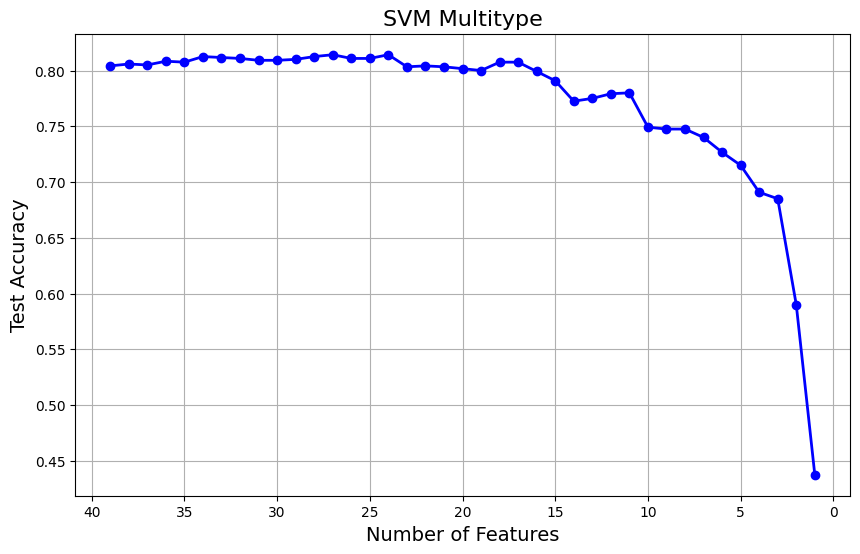}
    \\
    \adjustbox{scale=0.90}{(c) SVM binary}
    &
    \adjustbox{scale=0.90}{(d) SVM multiclass}
    \\ \\[-1.25ex]
    \includegraphics[width=0.45\textwidth]{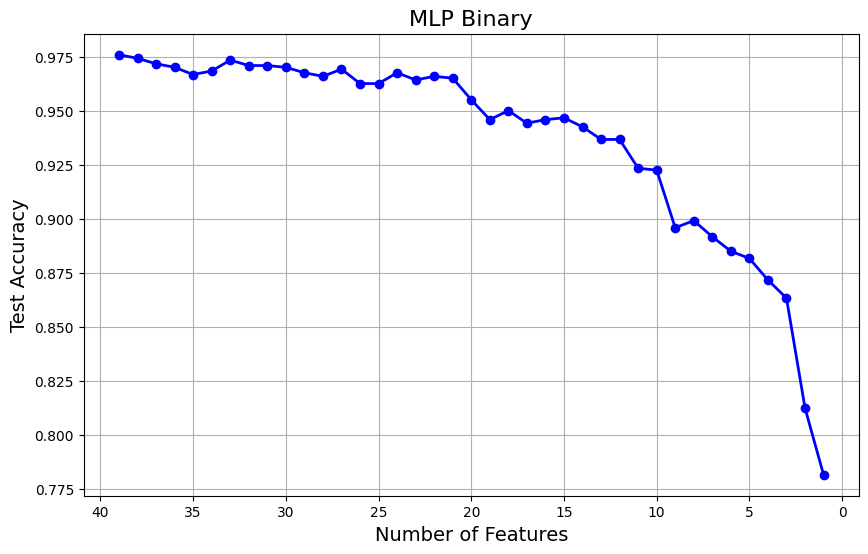}
    &
    \includegraphics[width=0.45\textwidth]{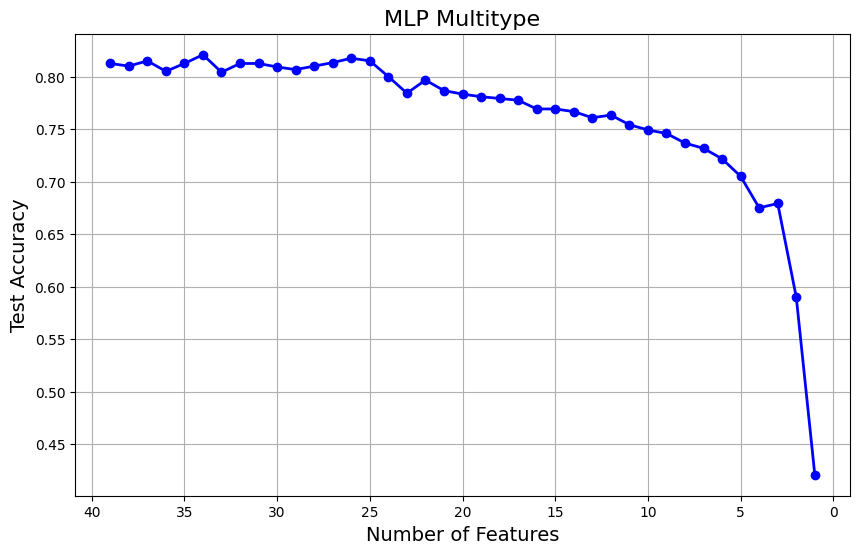}
    \\
    \adjustbox{scale=0.90}{(e) MLP binary}
    &
    \adjustbox{scale=0.90}{(f) MLP multiclass}
    \end{tabular}
    \caption{RFE results}\label{fig:RFE}
\end{figure}

From Figure~\ref{fig:RFE}(a) we observe that the highest binary classification 
accuracy of 0.9283 for LR is obtained using~36 features. However, we can achieve 
a binary classification accuracy of~0.9150 with just~20 features. As for multiclass classification, 
in Figure~\ref{fig:RFE}(b) we see that the best accuracy is obtained
using all~39 features, while in this case, an accuracy of~0.7492 is obtained 
when~22 features are used. 

\subsubsection{Reduced Feature SVM Results}

For the SVM model, Figure~\ref{fig:RFE}(c) shows that
the highest binary classification accuracy is~0.9792,
which is achieved with~33 or~34 features. 
It is also noteworthy that the binary classification accuracy is~0.9 
with just~9 features. Figure~\ref{fig:RFE}(d) shows that for 
multiclass classification, the highest accuracy is~0.8142,
and that this is achieved when training models with~24 features.
Furthermore, in the multiclass case, 
the accuracy is above~0.8 whenever the number of 
features is greater than~16. 

\subsubsection{Reduced Feature MLP Results}

From Figure~\ref{fig:RFE}(e), we see that the highest binary classification accuracy 
for the MLP model is achieved using~39 features. 
It is notable that the accuracy for binary classification is~0.9225 when 
using only~10 features, and it is more than~0.95 with~20 features.
Based on the results in Figure~\ref{fig:RFE}(f), we see that
the best multiclass classification accuracy is~0.8208,
which is achieved when training on~34 features.
For the multiclass case, an accuracy in excess of~0.8 is attained 
when using~24 features. 

\subsection{Convolutional Neural Network Results}

We train CNN models for both binary and multiclass classification 
based on the same image dataset used for the experiments above. 
However, instead of training on extracted features, these CNN
models are trained directly on the images.
We use a \texttt{Sequential} model from the Keras~\cite{chollet2015} library to train our CNN models.

We test the hyperparameters in Table~\ref{cnnResult},
where the best for the binary classification problem 
are given in \textbf{bold} font, the best performing for the multiclass 
case appear in \textit{italics}, and any that are best for
both cases are given in \textit{\textbf{italicized boldface}}.

\begin{table}[!htb]
\caption{CNN hyperparameters tested}\label{cnnResult}
\centering
\adjustbox{scale=0.90}{
\begin{tabular}{c|c} 
\midrule
\midrule
Hyperparameter & Values tested \\
\midrule
 layers & 6, 7, 8, \textit{9}, 10, \textbf{11} \\ 
 activation & softmax, \textit{\textbf{sigmoid}} \\
 dropout rate & \textbf{0.1}, 0.2, 0.3, 0.4, \textit{0.5} \\
 optimization & \textit{\textbf{Adam}} \\
 learning rate & \textit{\textbf{0.001}} \\ 
\midrule\midrule
\end{tabular}
}
\end{table}

For the binary classification problem, we obtain the best accuracy with a 
model consisting of one \texttt{Rescaling} layer, three \texttt{Conv2D} layers, 
three \texttt{MaxPooling2D} layers, one \texttt{Dropout} layer, one \texttt{Flatten} layer, 
and two \texttt{Dense} layers. The accuracy and loss graphs for this
case are given in Figures~\ref{fig:cnn_al}(a) and~(b) in the Appendix, respectively,
From these accuracy and loss graphs, 
we note that the best results are obtained with four training epochs,
after which overfitting is observed.

For multiclass classification, we obtain the best accuracy with one \texttt{Rescaling} layer, 
two \texttt{Conv2D} layers (with L2 regularization), two \texttt{MaxPooling2D} layers, 
one  \texttt{Dropout} layer, one \texttt{Flatten} layer, 
and two \texttt{Dense} layers, with the first dense layer employing L2 regularization. 
The accuracy and loss graphs for this
case are given in Figures~\ref{fig:cnn_al}(c) and~(d) in the Appendix, respectively.
From the accuracy and loss graphs for this multiclass problem, 
we note that the best results are obtained with~18 training epochs.

The validation accuracy of our best performing binary 
classification model is~0.9500, which is competitive with 
our best performing models. In contrast, our best CNN result 
for the multiclass case is~0.7550, which is the worst of all of the
four models considered.

\subsection{Discussion}

\begin{figure}[!htb]
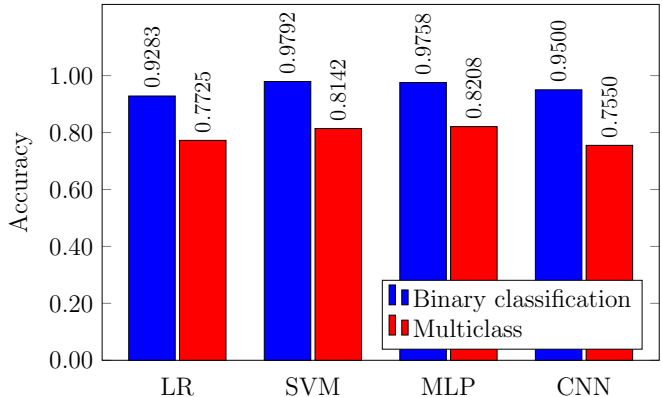

    \centering
    \input figures/bar_accuracy.tex
    \caption{Accuracy comparison}\label{fig:accuracy}   
\end{figure}

In Figure~\ref{fig:accuracy} we summarize the performance of the various models 
considered. Note that the best results for the LR model on multitype 
classification and MLP model on binary classification were obtained
using all features, while the best results for other experiments were obtained
using reduced feature sets.

We observe that the highest accuracy for binary classification in~0.9792,
which is achieved with the SVM model (after applying RFE). In addition,
the MLP model performs almost as well as the SVM model for binary classification. 
For the multiclass case, the MLP model achieves the highest 
accuracy at~0.8208, while the SVM performs nearly as well, 
with LR and CNN are significantly worse.

Confusion matrices for the multiclass LR, SVM, MLP, and CNN models
are given, respectively, in Figures~\ref{fig:conf}(a) through~(d)
in the Appendix. 
Interestingly, we observe that distinguishing between the
human-generated classes is the primary source of errors for all models, 
with the AI-generated samples being relatively 
easy to distinguish---both from human-generated art,
and from the other styles of AI-generated art.

\section{Conclusion}\label{sect:conclusion}

In this paper, we considered the problem of distinguish between  
human-generated and AI-generated art. In the binary classification case,
we achieved an accuracy in excess of~97\%. In the more challenging
multiclass case, where we classify each sample into one of the six categories
in our dataset, we achieved an accuracy of more than~82\%. In this latter case,
discriminating between the three classes of human-generated art proved
to be the greatest challenge.

Our best binary classification model was an SVM trained on a subset of~33 out
of the~39 features that we extracted. For the multiclass case, an MLP model
trained on~34 features was the most effective approach.
Overall, our results clearly indicate that learning models are highly effective 
at distinguishing between human-generated and AI-generated art, although 
they are considerably less effective at distinguishing between styles
of human-generated art.

For future work, additional styles of art, additional features, and other types of
models would all be worth considering. Given that the binary classification problem 
is relatively easy, it might be most effective to consider a two-stage approach, where
samples are first classified as human-generated or AI-generated (with high confidence),
then classified into a specific style of art (with lower confidence).

\bibliographystyle{alpha}
\bibliography{reference}

\section*{Appendix}
\renewcommand{\thesubsection}{A.\arabic{subsection}}
\setcounter{table}{0}
\renewcommand{\thetable}{A.\arabic{table}}
\setcounter{figure}{0}
\renewcommand{\thefigure}{A.\arabic{figure}}

In this appendix, we provide graphs of histograms for selected features.
We also provide accuracy and loss graphs for our CNN models
and confusion matrices for each of our models
in the multiclass case.

\begin{figure}[!htb]
    \centering
    \begin{tabular}{cc}
    \includegraphics[width=0.45\textwidth]{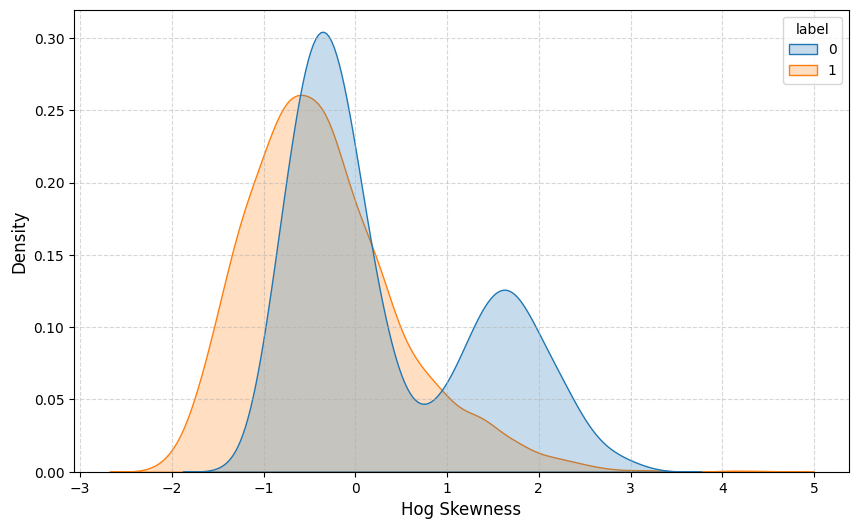}
    &
    \includegraphics[width=0.45\textwidth]{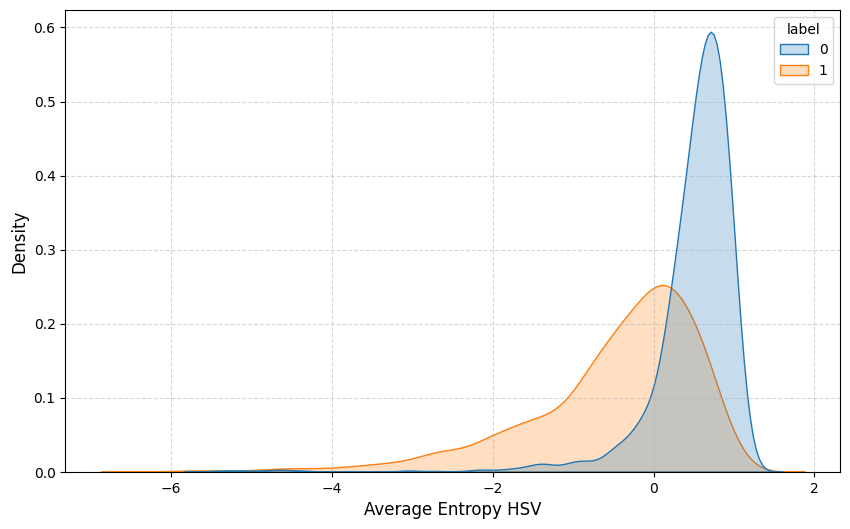} 
    \\
    \adjustbox{scale=0.90}{(a) Hog skewness}
    &
    \adjustbox{scale=0.90}{(b) Average entropy HSV}
    \\ \\[-1.25ex]
    \includegraphics[width=0.45\textwidth]{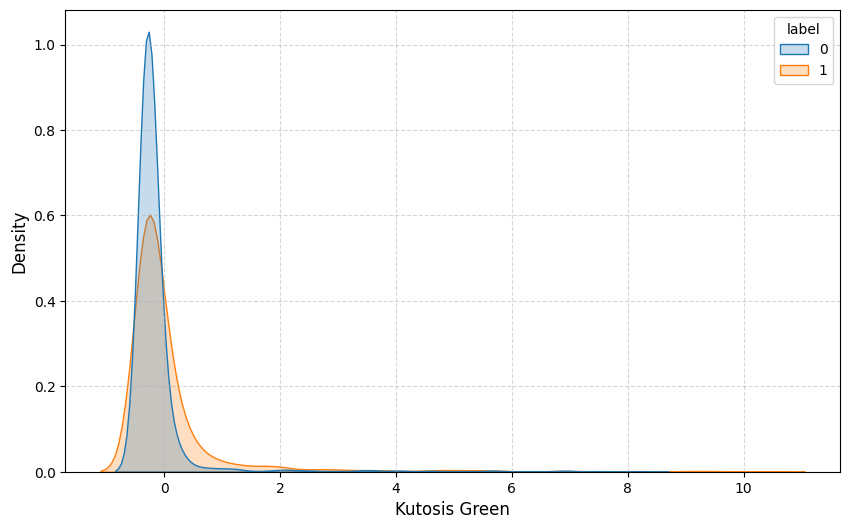}
    &
    \includegraphics[width=0.45\textwidth]{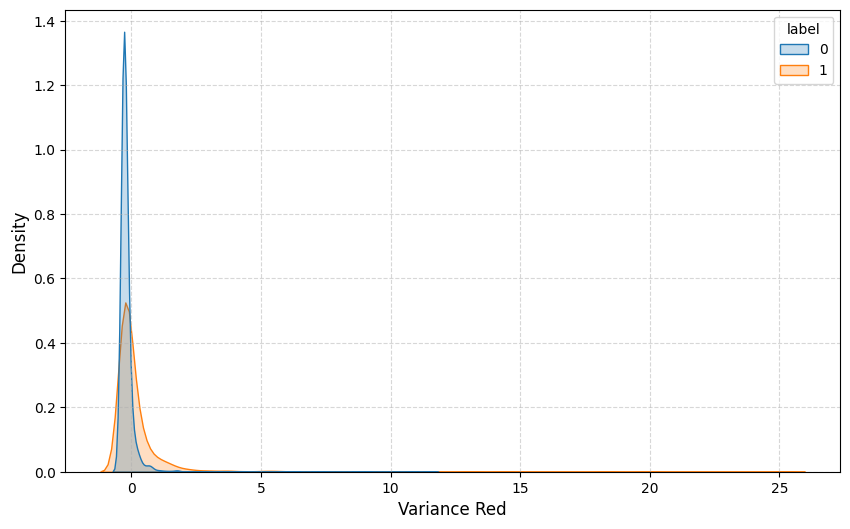}
    \\
    \adjustbox{scale=0.90}{(c) Green kurtosis}
    &
    \adjustbox{scale=0.90}{(d) Red variance}
    \end{tabular}
    \caption{Histograms for selected features}\label{fig:histos}
\end{figure}

\begin{figure}[!htb]
    \centering
    \begin{tabular}{cc}
    \includegraphics[width=0.46\textwidth]{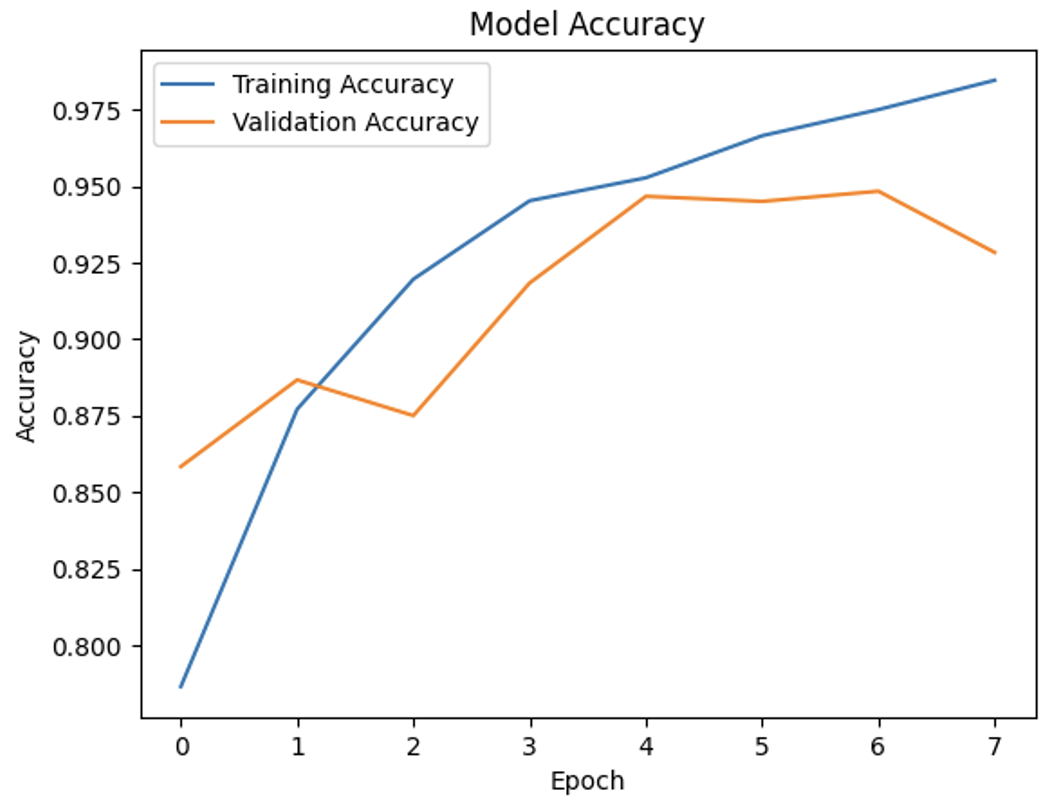}
    &
    \includegraphics[width=0.445\textwidth]{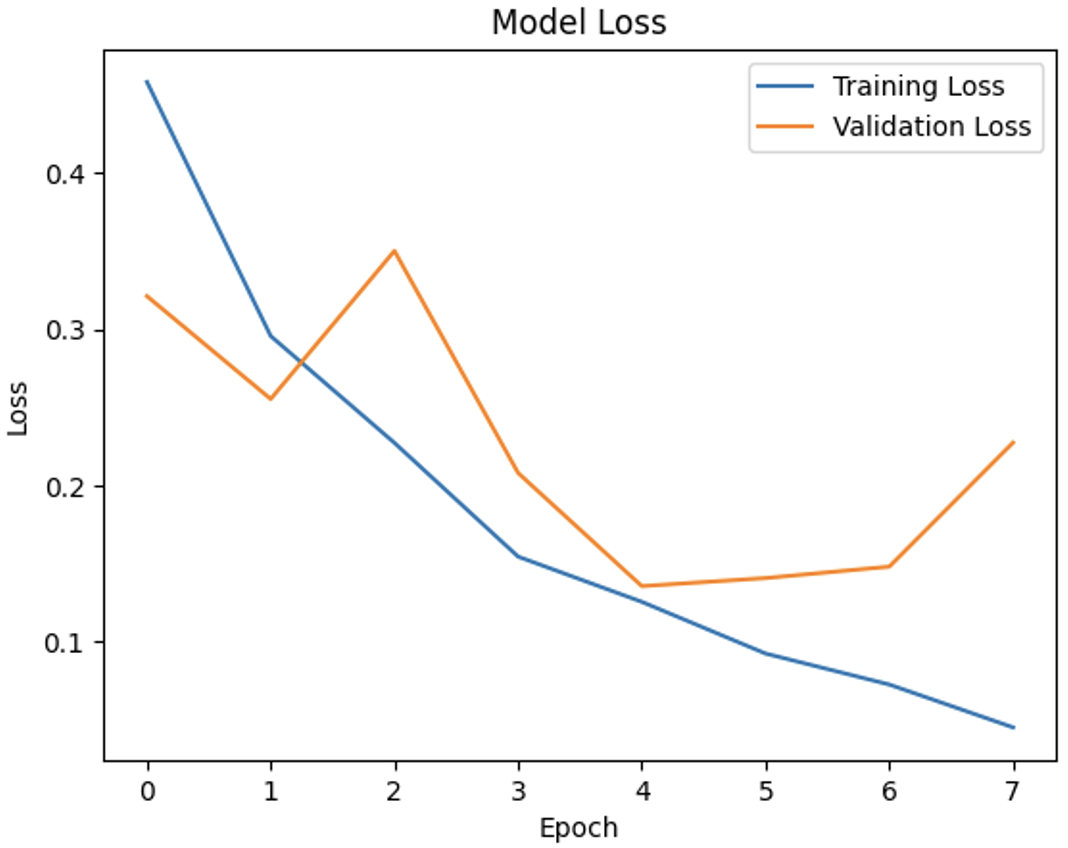} 
    \\
    \adjustbox{scale=0.90}{(a) Binary accuracy}
    &
    \adjustbox{scale=0.90}{(b) Binary loss}
    \\ \\[-1.25ex]
    \includegraphics[width=0.45\textwidth]{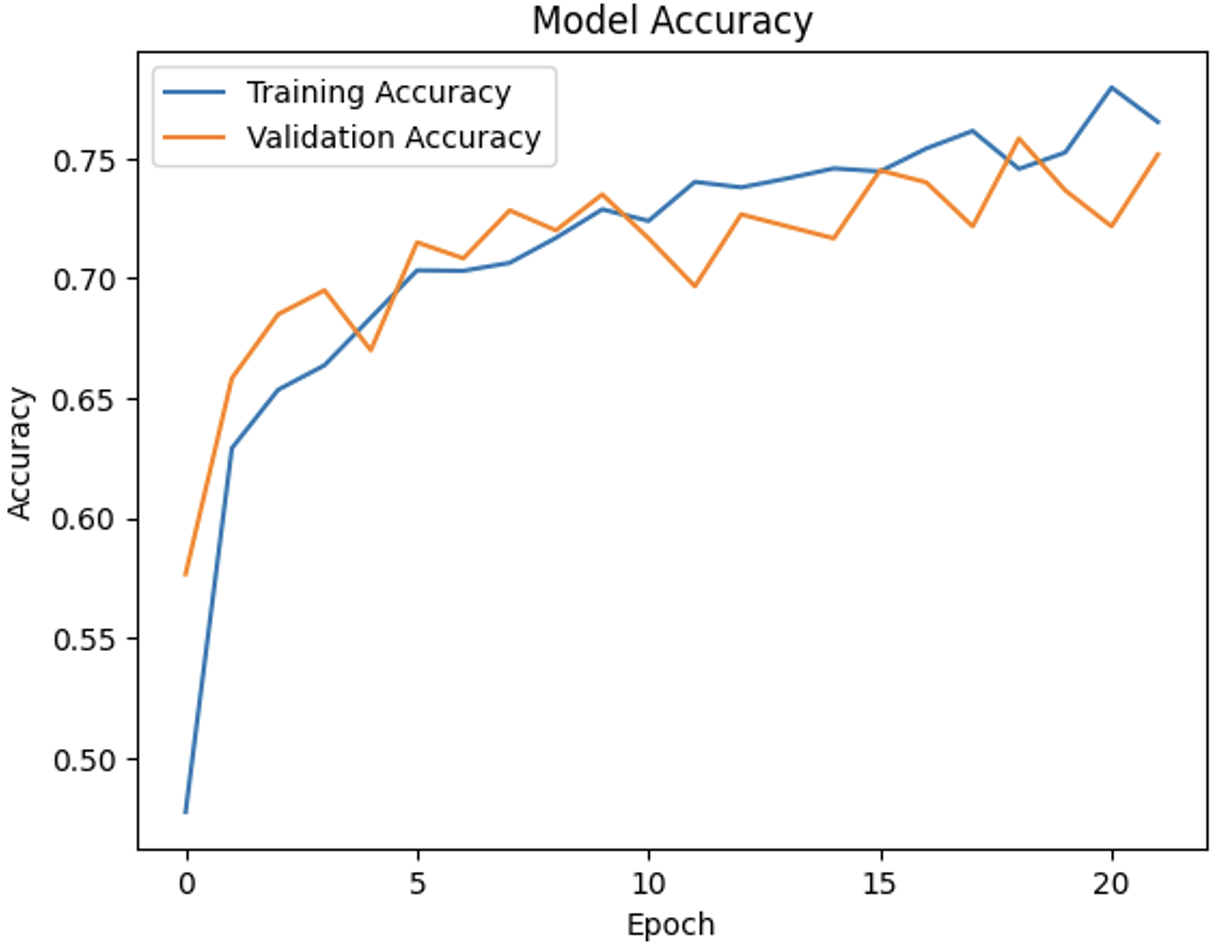}
    &
    \includegraphics[width=0.45\textwidth]{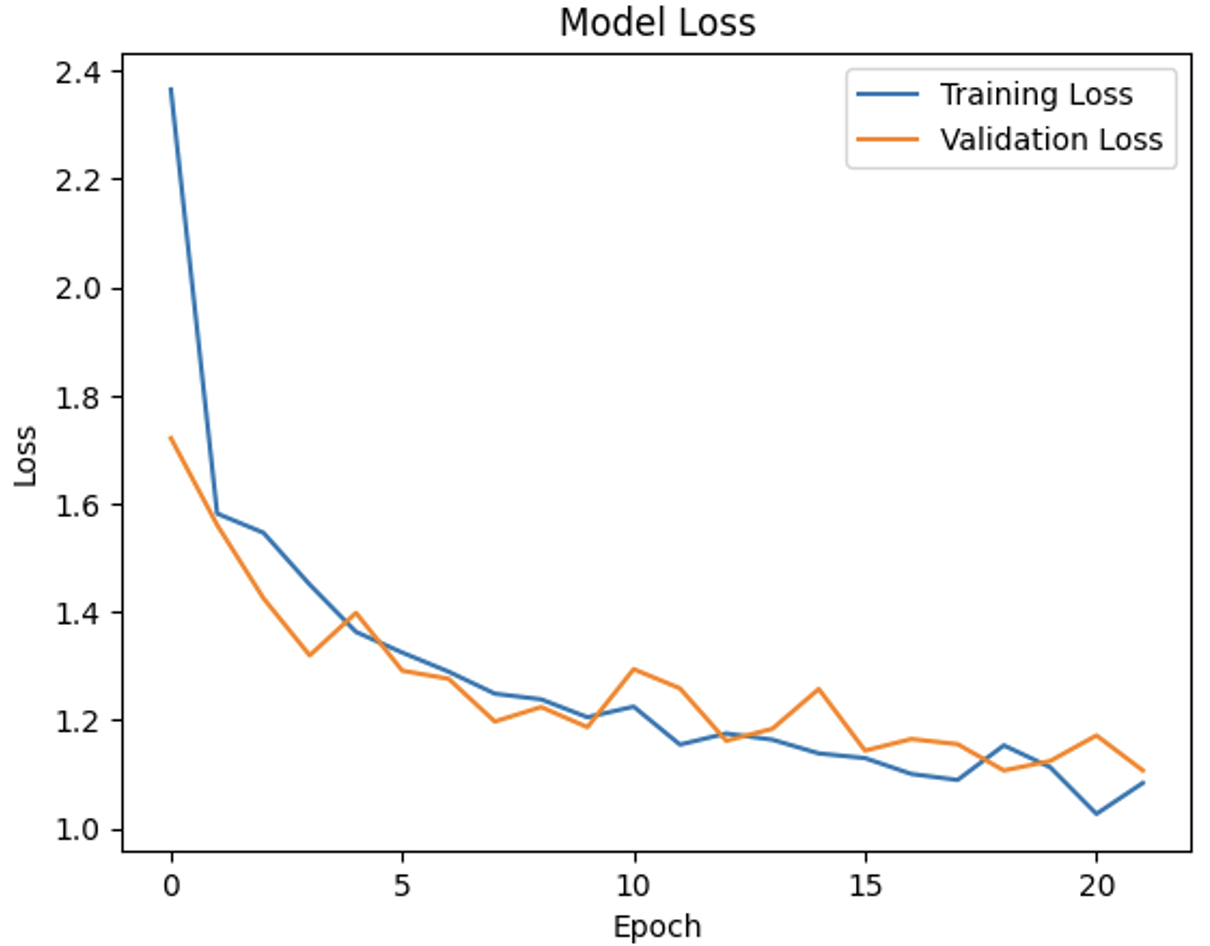}
    \\
    \adjustbox{scale=0.90}{(c) Multiclass accuracy}
    &
    \adjustbox{scale=0.90}{(d) Multiclass loss}
    \end{tabular}
    \caption{CNN accuracy and loss graphs}\label{fig:cnn_al}
\end{figure}

\begin{figure}[!htb]
    \centering
    \setlength{\tabcolsep}{-2.5pt}
    \begin{tabular}{cc}
    \adjustbox{scale=0.80}{
    \input figures/LR_conf.tex
    }
    &
    \adjustbox{scale=0.80}{
    \input figures/SVM_conf.tex
    }
    \\[-0.5ex]
    \adjustbox{scale=0.90}{(a) LR}
    &
    \adjustbox{scale=0.90}{(b) SVM}
    \\ \\[-1.25ex]
    \adjustbox{scale=0.80}{
    \input figures/MLP_conf.tex
    }
    &
    \adjustbox{scale=0.80}{
    \input figures/CNN_conf.tex
    }
    \\[-0.5ex]
    \adjustbox{scale=0.90}{(c) MLP}
    &
    \adjustbox{scale=0.90}{(d) CNN}
    \end{tabular}
    \caption{Mulitclass confusion matrix}\label{fig:conf}
\end{figure}

\end{document}

%% file: preamble.tex
\usepackage[english]{babel}

\usepackage[bb=boondox,bbscaled=1.0,cal=boondoxo]{mathalfa}

\advance\oddsidemargin by -0.35in
\advance\textwidth by 0.7in

\advance\topmargin by -0.4in
\advance\textheight by 0.8in

\usepackage{amsmath}
\usepackage{graphicx}
\usepackage[colorlinks=true, allcolors=blue]{hyperref}
\usepackage{multirow}
\usepackage{subfigure}
\usepackage{booktabs}
\usepackage{adjustbox}

\usepackage{pgfplots}
\pgfplotsset{compat=1.3} 
\usepackage{pgfplotstable}

\long\def\symbolfootnotetext[#1]#2{\begingroup%
\def\thefootnote{\fnsymbol{footnote}}\footnotetext[#1]{#2}\endgroup}

%% file: figures/bar_accuracy.tex
\begin{tikzpicture}[scale=0.8, every node/.style={scale=1.0}]
\pgfkeys{/pgf/number format/.cd,1000 sep={}}
\begin{axis}[
        width  = 0.7*\textwidth,
        height = 7.5cm,
        ymin=0.0,ymax=1.25,
        ytick={0.0, 0.2, 0.4, 0.6, 0.8, 1.0},
        major x tick style = transparent,
        ybar=5*\pgflinewidth,
        bar width=22pt,
        ylabel = {Accuracy},
        ylabel style = {scale = 1.0},
        symbolic x coords={LR, SVM, MLP, CNN},
        xticklabels={LR, SVM, MLP, CNN},
	y tick label style={
    		/pgf/number format/.cd,
   		fixed,
   		fixed zerofill,
    		precision=2},
        xtick = data,
        x tick label style={scale=1.0,
		},
        nodes near coords,
        every node near coord/.append style={rotate=90, scale=0.90,color=black,opacity=1.00,
        								   anchor=west, 
								   /pgf/number format/.cd,
								   fixed,
								   fixed zerofill,
								   precision=4},
        enlarge x limits=0.185,
        legend cell align=left,
        legend pos=south east,
]
\addplot [fill=blue,opacity=1.00]
coordinates {
(LR, 0.9283) 
(SVM, 0.9792) 
(MLP, 0.9758) 
(CNN, 0.9500)
};
\addlegendentry{Binary classification}
\addplot [fill=red,opacity=1.00]
coordinates {
(LR, 0.7725) 
(SVM, 0.8142) 
(MLP, 0.8208) 
(CNN, 0.7550)
};
\addlegendentry{Multiclass}
\end{axis}
\end{tikzpicture}

%% file: figures/LR_conf.tex
\begin{tikzpicture}[scale=0.4]
    \begin{axis}[
        width=15cm,
        height=15cm,
	colormap={bluewhite}{color=(white) rgb255=(100,149,237)},
        xticklabels={
AI baroque,
AI cubism,
AI exp.,
Human baroque,
Human cubism,
Human exp.
        },
        xtick={0,...,5},
        xtick style={draw=none},
	xticklabel style={anchor=east,rotate=60,yshift=-5pt,scale=1.75},
        yticklabels={
AI baroque,
AI cubism,
AI exp.,
Human baroque,
Human cubism,
Human exp.
        },
        ytick={0,...,5},
        ytick style={draw=none},
        enlargelimits=false,
        yticklabel style={scale=1.75},
        colorbar,
        colorbar style={
            ytick={0,40,80,120,160,200},
            yticklabels={0,40,80,120,160,200},
            yticklabel={\pgfmathprintnumber\tick},
            yticklabel style={
            		scale=1.75,
            		/pgf/number format/fixed,
			/pgf/number format/fixed zerofill,
			/pgf/number format/precision=0}
        },
        point meta min=0,
        point meta max=200,
        nodes near coords={\pgfmathprintnumber\pgfplotspointmeta},
        nodes near coords black white/.style={
            small value/.style={
                yshift=-12pt,
                text=black,
                /pgf/number format/fixed,
                /pgf/number format/precision=0,
                /pgf/number format/zerofill=true,
                scale=1.75,
            },
            large value/.style={
                yshift=-12pt,
                text=white,
                /pgf/number format/fixed,
                /pgf/number format/precision=0,
                /pgf/number format/zerofill=true,
                scale=1.75,
            },
            every node near coord/.style={
                check for zero/.code={
                    \pgfmathfloatifflags{\pgfplotspointmeta}{0}{
                        \pgfkeys{/tikz/coordinate}
                    }{
                        \begingroup
                        \pgfkeys{/pgf/fpu}
                        \pgfmathparse{\pgfplotspointmeta<#1}
                        \global\let\result=\pgfmathresult
                        \endgroup
                        %
                        %
                        \pgfmathfloatcreate{1}{1.0}{0}
                        \let\ONE=\pgfmathresult
                        \ifx\result\ONE
                            \pgfkeysalso{/pgfplots/small value}
                        \else
                            \pgfkeysalso{/pgfplots/large value}
                        \fi
                    }
                },
                check for zero,
            },
        },
        nodes near coords black white=100,
    ]
        \addplot[
            matrix plot,
            mesh/cols=6,
            point meta=explicit,draw=gray
        ] table [meta=C] {
            x y C
0 0 190
1 0 0
2 0 1
3 0 2
4 0 2
5 0 5
0 1 3
1 1 181
2 1 3
3 1 0
4 1 12
5 1 1
0 2 0
1 2 0
2 2 199
3 2 0
4 2 1
5 2 0
0 3 13
1 3 0
2 3 0
3 3 143
4 3 11
5 3 33
0 4 11
1 4 13
2 4 0
3 4 15
4 4 101
5 4 60
0 5 10
1 5 1
2 5 1
3 5 49
4 5 40
5 5 99
         };
    \end{axis}
\end{tikzpicture}

%% file: figures/SVM_conf.tex
\begin{tikzpicture}[scale=0.4]
    \begin{axis}[
        width=15cm,
        height=15cm,
	colormap={bluewhite}{color=(white) rgb255=(100,149,237)},
        xticklabels={
AI baroque,
AI cubism,
AI exp.,
Human baroque,
Human cubism,
Human exp.
        },
        xtick={0,...,5},
        xtick style={draw=none},
	xticklabel style={anchor=east,rotate=60,yshift=-5pt,scale=1.75},
        yticklabels={
AI baroque,
AI cubism,
AI exp.,
Human baroque,
Human cubism,
Human exp..
        },
        ytick={0,...,5},
        ytick style={draw=none},
        enlargelimits=false,
        yticklabel style={scale=1.75},
        colorbar,
        colorbar style={
            ytick={0,40,80,120,160,200},
            yticklabels={0,40,80,120,160,200},
            yticklabel={\pgfmathprintnumber\tick},
            yticklabel style={
            		scale=1.75,
            		/pgf/number format/fixed,
			/pgf/number format/fixed zerofill,
			/pgf/number format/precision=0}
        },
        point meta min=0,
        point meta max=200,
        nodes near coords={\pgfmathprintnumber\pgfplotspointmeta},
        nodes near coords black white/.style={
            small value/.style={
                yshift=-12pt,
                text=black,
                /pgf/number format/fixed,
                /pgf/number format/precision=0,
                /pgf/number format/zerofill=true,
                scale=1.75,
            },
            large value/.style={
                yshift=-12pt,
                text=white,
                /pgf/number format/fixed,
                /pgf/number format/precision=0,
                /pgf/number format/zerofill=true,
                scale=1.75,
            },
            every node near coord/.style={
                check for zero/.code={
                    \pgfmathfloatifflags{\pgfplotspointmeta}{0}{
                        \pgfkeys{/tikz/coordinate}
                    }{
                        \begingroup
                        \pgfkeys{/pgf/fpu}
                        \pgfmathparse{\pgfplotspointmeta<#1}
                        \global\let\result=\pgfmathresult
                        \endgroup
                        %
                        %
                        \pgfmathfloatcreate{1}{1.0}{0}
                        \let\ONE=\pgfmathresult
                        \ifx\result\ONE
                            \pgfkeysalso{/pgfplots/small value}
                        \else
                            \pgfkeysalso{/pgfplots/large value}
                        \fi
                    }
                },
                check for zero,
            },
        },
        nodes near coords black white=100,
    ]
        \addplot[
            matrix plot,
            mesh/cols=6,
            point meta=explicit,draw=gray
        ] table [meta=C] {
            x y C
0 0 194
1 0 0
2 0 0
3 0 1
4 0 4
5 0 1
0 1 0
1 1 195
2 1 0
3 1 0
4 1 5
5 1 0
0 2 0
1 2 1
2 2 198
3 2 0
4 2 1
5 2 0
0 3 12
1 3 0
2 3 0
3 3 152
4 3 10
5 3 26
0 4 9
1 4 5
2 4 0
3 4 14
4 4 114
5 4 58
0 5 5
1 5 0
2 5 0
3 5 39
4 5 44
5 5 112
         };
    \end{axis}
\end{tikzpicture}

%% file: figures/MLP_conf.tex
\begin{tikzpicture}[scale=0.4]
    \begin{axis}[
        width=15cm,
        height=15cm,
	colormap={bluewhite}{color=(white) rgb255=(100,149,237)},
        xticklabels={
AI baroque,
AI cubism,
AI exp.,
Human baroque,
Human cubism,
Human exp.
        },
        xtick={0,...,5},
        xtick style={draw=none},
	xticklabel style={anchor=east,rotate=60,yshift=-5pt,scale=1.75},
        yticklabels={
AI baroque,
AI cubism,
AI exp.,
Human baroque,
Human cubism,
Human exp.
        },
        ytick={0,...,5},
        ytick style={draw=none},
        enlargelimits=false,
        yticklabel style={scale=1.75},
        colorbar,
        colorbar style={
            ytick={0,40,80,120,160,200},
            yticklabels={0,40,80,120,160,200},
            yticklabel={\pgfmathprintnumber\tick},
            yticklabel style={
            		scale=1.75,
            		/pgf/number format/fixed,
			/pgf/number format/fixed zerofill,
			/pgf/number format/precision=0}
        },
        point meta min=0,
        point meta max=200,
        nodes near coords={\pgfmathprintnumber\pgfplotspointmeta},
        nodes near coords black white/.style={
            small value/.style={
                yshift=-12pt,
                text=black,
                /pgf/number format/fixed,
                /pgf/number format/precision=0,
                /pgf/number format/zerofill=true,
                scale=1.75,
            },
            large value/.style={
                yshift=-12pt,
                text=white,
                /pgf/number format/fixed,
                /pgf/number format/precision=0,
                /pgf/number format/zerofill=true,
                scale=1.75,
            },
            every node near coord/.style={
                check for zero/.code={
                    \pgfmathfloatifflags{\pgfplotspointmeta}{0}{
                        \pgfkeys{/tikz/coordinate}
                    }{
                        \begingroup
                        \pgfkeys{/pgf/fpu}
                        \pgfmathparse{\pgfplotspointmeta<#1}
                        \global\let\result=\pgfmathresult
                        \endgroup
                        %
                        %
                        \pgfmathfloatcreate{1}{1.0}{0}
                        \let\ONE=\pgfmathresult
                        \ifx\result\ONE
                            \pgfkeysalso{/pgfplots/small value}
                        \else
                            \pgfkeysalso{/pgfplots/large value}
                        \fi
                    }
                },
                check for zero,
            },
        },
        nodes near coords black white=100,
    ]
        \addplot[
            matrix plot,
            mesh/cols=6,
            point meta=explicit,draw=gray
        ] table [meta=C] {
            x y C
0 0 192
1 0 0
2 0 0
3 0 2
4 0 5
5 0 1
0 1 0
1 1 192
2 1 0
3 1 0
4 1 8
5 1 0
0 2 0
1 2 0
2 2 199
3 2 0
4 2 1
5 2 0
0 3 4
1 3 0
2 3 0
3 3 152
4 3 10
5 3 34
0 4 8
1 4 5
2 4 0
3 4 10
4 4 124
5 4 53
0 5 5
1 5 1
2 5 0
3 5 35
4 5 43
5 5 116
         };
    \end{axis}
\end{tikzpicture}

%% file: figures/CNN_conf.tex
\begin{tikzpicture}[scale=0.4]
    \begin{axis}[
        width=15cm,
        height=15cm,
	colormap={bluewhite}{color=(white) rgb255=(100,149,237)},
        xticklabels={
AI baroque,
AI cubism,
AI exp.,
Human baroque,
Human cubism,
Human exp.
        },
        xtick={0,...,5},
        xtick style={draw=none},
	xticklabel style={anchor=east,rotate=60,yshift=-5pt,scale=1.75},
        yticklabels={
AI baroque,
AI cubism,
AI exp.,
Human baroque,
Human cubism,
Human exp.
        },
        ytick={0,...,5},
        ytick style={draw=none},
        enlargelimits=false,
        yticklabel style={scale=1.75},
        colorbar,
        colorbar style={
            ytick={0,20,40,60,80,100},
            yticklabels={0,20,40,60,80,100},
            yticklabel={\pgfmathprintnumber\tick},
            yticklabel style={
            		scale=1.75,
            		/pgf/number format/fixed,
			/pgf/number format/fixed zerofill,
			/pgf/number format/precision=0}
        },
        point meta min=0,
        point meta max=100,
        nodes near coords={\pgfmathprintnumber\pgfplotspointmeta},
        nodes near coords black white/.style={
            small value/.style={
                yshift=-12pt,
                text=black,
                /pgf/number format/fixed,
                /pgf/number format/precision=0,
                /pgf/number format/zerofill=true,
                scale=1.75,
            },
            large value/.style={
                yshift=-12pt,
                text=white,
                /pgf/number format/fixed,
                /pgf/number format/precision=0,
                /pgf/number format/zerofill=true,
                scale=1.75,
            },
            every node near coord/.style={
                check for zero/.code={
                    \pgfmathfloatifflags{\pgfplotspointmeta}{0}{
                        \pgfkeys{/tikz/coordinate}
                    }{
                        \begingroup
                        \pgfkeys{/pgf/fpu}
                        \pgfmathparse{\pgfplotspointmeta<#1}
                        \global\let\result=\pgfmathresult
                        \endgroup
                        %
                        %
                        \pgfmathfloatcreate{1}{1.0}{0}
                        \let\ONE=\pgfmathresult
                        \ifx\result\ONE
                            \pgfkeysalso{/pgfplots/small value}
                        \else
                            \pgfkeysalso{/pgfplots/large value}
                        \fi
                    }
                },
                check for zero,
            },
        },
        nodes near coords black white=50,
    ]
        \addplot[
            matrix plot,
            mesh/cols=6,
            point meta=explicit,draw=gray
        ] table [meta=C] {
            x y C
0 0 93
1 0 0
2 0 0
3 0 4
4 0 2
5 0 1
0 1 0
1 1 92
2 1 2
3 1 0
4 1 6
5 1 0
0 2 0
1 2 2
2 2 98
3 2 0
4 2 0
5 2 0
0 3 5
1 3 0
2 3 0
3 3 83
4 3 5
5 3 7
0 4 1
1 4 2
2 4 0
3 4 19
4 4 54
5 4 24
0 5 2
1 5 0
2 5 0
3 5 40
4 5 19
5 5 39
         };
    \end{axis}
\end{tikzpicture}